\documentclass[letterpaper, 10 pt, conference]{ieeeconf}  % Comment
\IEEEoverridecommandlockouts                              % This
\usepackage{amsmath} % assumes amsmath package installed
\usepackage{amssymb}  % assumes amsmath package installed

\usepackage[patch={footnote,item,verbatim}]{microtype}
\usepackage{booktabs,multirow}
\usepackage{graphicx}
\makeatletter
\def\endfigure{\end@float}\def\endtable{\end@float}
\makeatother
\usepackage[caption=false]{subfig}
\usepackage{url}

\title{\LARGE \bf
  DetAug: Obstacle-Blind Trajectory Augmentation \\
  for Zero-shot Obstacle Avoidance
}

\author{Reece O'Mahoney, Moritz Zoellner and Ioannis Havoutis% <-this
  \thanks{Reece O'Mahoney and Ioannis Havoutis are with the Oxford Robotics
    Institute, University of Oxford, Oxford, UK\newline
  {\tt\small \{reeceo,ioannis\}@robots.ox.ac.uk}}%
  \thanks{Moritz Zoellner is with Purdue University, West Lafayette, IN, USA
  {\tt\small zoellner@purdue.edu}}%
}

\begin{document}

\maketitle
\thispagestyle{empty}
\pagestyle{empty}

%%%%%%%%%%%%%%%%%%%%%%%%%%%%%%%%%%%%%%%%%%%%%%%%%%%%%%%%%%%%%%%%%%%%%%%%%%%%%%%%
\begin{abstract}

  Policies for robotic manipulation are produced by training on large
  teleoperated datasets. These datasets typically consist of
  free-space trajectories, making them difficult to transfer to
  test-time environments with obstacles. Previous methods for closing
  this gap have largely fallen into two groups. Dataset augmentation
  addresses it at training time but needs obstacle geometry in advance, whereas
  steering an existing checkpoint at inference time avoids that
  requirement but is limited in flexibility. Our method draws from both
  areas without inheriting either drawback. DetAug applies
  an obstacle-blind augmentation scheme to the transit phases of a
  free-space dataset, leaving object interactions untouched, and
  records the augmentation parameters as an explicit conditioning
  label. At inference it samples a batch of labels and executes the
  trajectory with the lowest collision cost. On the SafeLIBERO
  benchmark DetAug achieves a collision-free success rate more than
  20pp above the next best method, and selecting over the label space
  outperforms guidance on the same policy by 26pp. On real hardware,
  inference-time steering methods collapse on tasks requiring large
  detours, while DetAug matches or exceeds an obstacle-conditioned
  baseline without ever seeing obstacles in training.

\end{abstract}

%%%%%%%%%%%%%%%%%%%%%%%%%%%%%%%%%%%%%%%%%%%%%%%%%%%%%%%%%%%%%%%%%%%%%%%%%%%%%%%%
\section{INTRODUCTION}

Manipulation policies are most often trained from demonstrations
collected in free-space \cite{zhao2023aloha, chi2023diffusion,
oxe2024, khazatsky2024droid}, but real deployment environments
frequently contain obstacles that are absent in the dataset,
collapsing performance \cite{yang2025cape, wang2026embodisteer}.
Previous methods for dealing with this in the
literature mainly fall into two camps.

The first consists of inference-time techniques that modify the
behaviour of a frozen policy. Since state-of-the-art policies are
typically diffusion \cite{chi2023diffusion} or flow matching
\cite{lipman2023flow, black2024pi0, bjorck2025gr00t} based, these
techniques normally use guidance \cite{dhariwal2021guidance,
janner2022diffuser} to steer the policy output via a gradient derived
from some collision signal \cite{carvalho2024mpd, saha2024edmp,
li2024lano3dp, song2026omniguide, wang2026embodisteer}.
The second involves synthetic data generation that samples a
distribution of obstacles at training time, and augments the dataset
with collision-free trajectories that are conditioned on the obstacle
geometry \cite{xue2025demogen}. While both of these approaches have
shown some success they also have notable drawbacks.

Inference-time methods are easy to apply to a pre-existing policy, and so
can inherit the capabilities of a model trained on web-scale data
\cite{brohan2023rt2, kim2024openvla, black2024pi0}, but suffer from
limited flexibility, as the possible behaviours are limited to what
already exists in the prior \cite{yang2025cape}. When significant
detours from the default behaviour are required, these methods do not
fail by colliding but by trading task completion for safety: the
steered policy stalls or times out rather than finding a path around
the obstacle. Selection-based variants that search over the prior's
own samples \cite{ramesh2025testtime} inherit the same ceiling, which
we measure directly in Section \ref{sec:ncond}. Train-time methods do
not have this weakness, but are instead limited by requiring the
obstacle geometry at training time: the detours they distil are fixed
by the obstacles sampled during data generation, and we find that at
a matched architecture and training budget, a policy conditioned on
the obstacle geometry makes only weak use of it, even when the
deployment obstacles are drawn from that same distribution
\cite{xue2025demogen}. The gap is widest when adapting a
pre-existing policy: the relatively heavy conditioning required to
capture 3D obstacle geometry is difficult to graft onto a frozen
backbone through a lightweight conditioning pathway such as
AdaLN-Zero \cite{peebles2023dit}, whereas a low-dimensional label is
not. Our
method, \textbf{Detour Augmentation (DetAug)}, builds on the strengths
of both of these families of techniques while avoiding their
weaknesses: steering methods assume the required modes already exist,
and augmentation methods assume the obstacles are known; we assume
neither.

\begin{figure}[t]
  \centering
  \includegraphics[width=0.9\columnwidth]{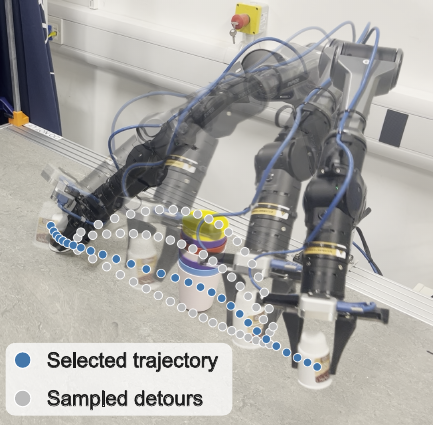}
  \caption{DetAug on hardware. A policy trained only on free-space
    demos augmented with random detours avoids an obstacle never
    seen in training. At each replan, a batch of detours (grey) is
    sampled from the label space and the collision-free candidate
  (blue) is executed.}
  \label{fig:teaser}
\end{figure}

\begin{figure*}[t]
  \centering
  \subfloat[Augment trajectories]{%
  \includegraphics[height=0.3\textwidth]{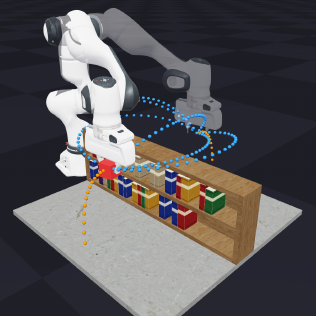}}\hfil
  \subfloat[Inject augmentation label]{%
  \includegraphics[height=0.3\textwidth]{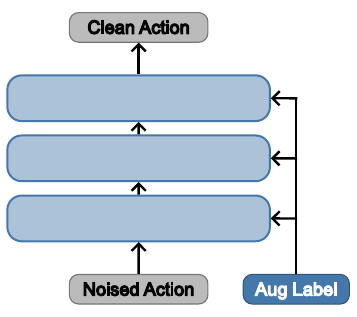}}\hfil
  \subfloat[Select at inference]{%
  \includegraphics[height=0.3\textwidth]{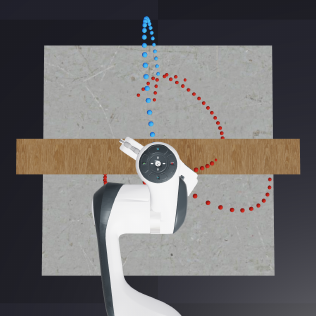}}
  \caption{Overview of DetAug. (a) Transit phases of free-space demos
    are replaced with random arcs, keeping grasp and release intact;
    the arc parameters form a low-dimensional label. (b) The label is
    injected into a flow-matching policy via AdaLN-Zero. (c) At
    inference, $K$ labels are sampled, each yields a trajectory, and
  the lowest collision-cost candidate (blue) is executed.}
  \label{fig:overview}
\end{figure*}

DetAug starts from a pre-collected teleoperation dataset and, following
DemoGen \cite{xue2025demogen}, segments each demonstration into
transit phases and interaction phases. Only the transit
phases are modified; the grasp and release windows are preserved
exactly. As a result, avoidance can never corrupt a grasp or release,
which is the dominant failure mode we observe on hardware for guidance
and constraint-based corrections that perturb the whole trajectory
(Section \ref{sec:hardware}). In particular, we randomly generate
arcs in Cartesian space, parameterised by the subtended angle and
orientation, and substitute them for the existing segments. The
parameters of the augmentation act as a conditioning variable that is
used during training. At inference, instead of doing a single forward
pass to generate an action, we batch these over a randomly sampled
set of conditioning values and check for collisions. A fresh batch is
drawn and re-scored at every replan. Section \ref{sec:method}
contains a more detailed explanation of this implementation.

By applying this method to a range of simulated and real tasks we
show the following contributions:

\begin{enumerate}

  \item Collision-free success on the SafeLIBERO benchmark more than
    20pp above the next best method, including an obstacle-conditioned
    policy trained at a matched architecture and budget, despite
    having no access to obstacle geometry during training.
  \item Superior task completion to inference-time methods in tasks
    requiring large deviations from the default behaviour, in
    simulation and on real hardware, where steering methods avoid the
    obstacle but fail to complete the task.
  \item Selection over the augmentation label space outperforms
    guidance on the same policy with the same collision cost by 26pp,
    and random labels dominate selection over the prior's own samples
    at every batch size.
  \item Far more effective compute-limited adaptation of a
    pre-trained policy than obstacle conditioning: with a frozen
    backbone and a 3M parameter adapter, DetAug reaches 35\% success
    versus 2\%.

\end{enumerate}

\section{RELATED WORK}

\subsection{Inference-Time Steering of Frozen Policies}

Guidance works by using the gradient of some differentiable cost
function to steer a diffusion process \cite{dhariwal2021guidance,
janner2022diffuser}. In the obstacle avoidance literature,
MPD \cite{carvalho2024mpd} guides a trajectory prior with collision
and smoothness costs, and EDMP \cite{saha2024edmp} guides sub-batches
of a scene-agnostic prior with an ensemble
of collision costs and keeps the lowest-cost candidate; it
diversifies by collision model and hyperparameter schedule where we
diversify by an explicit augmentation label. APEX
\cite{dastider2024apex} guides a
latent prior trained on collision-free trajectories conditioned on
the obstacle position. For visuomotor policies, Lan-o3dp
\cite{li2024lano3dp}, VLS \cite{liu2026vls} and OmniGuide
\cite{song2026omniguide} derive the steering signal from geometric
costs, VLM-synthesised rewards and attractor/repeller energy fields
respectively. EmbodiSteer \cite{wang2026embodisteer} applies a
CBF-QP correction in joint space at each denoising step, and AEGIS
\cite{hu2025vlsa} solves a similar CBF-QP once on the final action of
a frozen VLA, modelling the end effector and obstacles as ellipsoids
in Cartesian space. ReGuide
\cite{lin2026reguide} steers with a learned dynamics model and
recycles the guided rollouts into further training. Selection-based
variants search over the noise trajectory and keep the highest-reward
candidate at each step \cite{ramesh2025testtime}, and PDP
\cite{zhang2026pdp} fits a
behaviour latent learned from demonstrations to a new demonstration
by gradient descent, so the adapted behaviour stays within or near
the demonstrated space.

All of these can only reweight, deflect or choose among modes the
frozen prior already holds. CAPE \cite{yang2025cape} names this
drawback explicitly, noting that guidance only works when the data already
yields rich multimodal coverage, and attempts to mitigate it
inference-side by re-noising and re-denoising the unexecuted
remainder of the plan under collision guidance.

\subsection{Train-Time Augmentation with Synthetic Obstacles}

MimicGen \cite{mandlekar2023mimicgen} splits demonstrations into
object-centric segments, rigidly transforms each to a new object
pose and bridges them with interpolated motion; DemoGen
\cite{xue2025demogen} replans the free-space segments with a motion
planner and, for obstacle avoidance, samples obstacle primitives,
inserts them into the point cloud and plans a detour, yielding an
obstacle-conditioned policy. GLIDE \cite{li2024glide} likewise
distils planner-generated trajectories into a policy, though without
scene obstacles. These methods produce genuinely new behaviour, but
avoidance is fixed at data-generation time by the sampled object
distribution, with no test-time adaptation. Dense geometric conditioning also
needs a heavier mechanism to graft onto a pre-trained model, such as
a trainable encoder copy joined by zero convolutions
\cite{zhang2023controlnet}. DetAug is \emph{obstacle-blind}: the
augmentation is parameterised by a low-dimensional arc descriptor
rather than scene geometry; as such, the conditioning is small enough
to add via a single zero-initialised modulation pathway \cite{peebles2023dit}.

\subsection{Trajectory Stitching and Splitting}

Recombining segments is also studied as stitching to widen state
coverage \cite{lee2025scots}. Previous work
\cite{omahoney2025stitching} re-noised and split the generated plan
into halves at inference, re-denoising each under signed-distance and
smoothness guidance; DetAug removes the splitting and moves the fix
to the data, using geometry only to rank candidates rather than
construct trajectories.

\begin{table*}[!t]
  \caption{Quantitative results on the SafeLIBERO benchmark,
  aggregated by suite.}
  \label{tab:sim_results}
  \centering
  \small
  \setlength{\tabcolsep}{4pt}
  \begin{tabular}{l|ccc|ccc|ccc|ccc|ccc}
    \toprule
    & \multicolumn{3}{c|}{Spatial} & \multicolumn{3}{c|}{Object}
    & \multicolumn{3}{c|}{Goal} & \multicolumn{3}{c|}{Long}
    & \multicolumn{3}{c}{\textbf{Average}}\\
    \cmidrule(lr){2-4}\cmidrule(lr){5-7}\cmidrule(lr){8-10}\cmidrule(lr){11-13}\cmidrule(lr){14-16}
    Method & Clean & Total & Avoid & Clean & Total & Avoid
    & Clean & Total & Avoid & Clean & Total & Avoid
    & Clean & Total & Avoid\\
    \midrule
    DetAug (ours) & \textbf{58.7} & \textbf{73.0} & 68.0 &
    \textbf{88.5} & \textbf{92.5} & \textbf{94.0} & \textbf{32.5} &
    \textbf{58.0} & 43.5
    & \textbf{29.1} & \textbf{46.9} & 56.0 & \textbf{52.2} &
    \textbf{67.6} & 65.4\\
    DemoGen~\cite{xue2025demogen} & 19.4 & 51.7 & 24.5 & 32.4 & 80.7
    & 33.5 & 22.5 & 56.8 & 24.8 & 7.0 & 30.2 & 22.4 & 20.3
    & 54.9 & 26.3\\
    CAPE~\cite{yang2025cape} & 42.0 & 46.7 & \textbf{71.2} & 52.2 & 57.1
    & 87.2 & 25.8 & 37.2 & \textbf{61.2} & 2.2 & 3.0 & \textbf{79.0} & 30.6
    & 36.0 & \textbf{74.7}\\
    AEGIS~\cite{hu2025vlsa} & 28.1 & 43.0 & 57.1 & 47.2 &
    60.6 & 67.2 & 17.2 & 24.2 & 59.3 & 13.8
    & 15.6 & 73.0 & 26.6 & 35.9 & 64.2\\
    \bottomrule
  \end{tabular}
  \par\smallskip
  \parbox{0.95\textwidth}{\footnotesize
    All entries are percentages, averaged over the Level I (obstacle
    near the target) and Level II (obstacle on the motion path)
    scenarios of each suite. \textit{Average} is the mean
    over the four suites. \textit{Clean} is the collision-free task
    success rate, \textit{Total} is the task success rate irrespective
    of collisions, and \textit{Avoid} is the collision avoidance rate.
  The best result in each column is shown in bold.}
\end{table*}

\section{METHOD}\label{sec:method}

\subsection{Overview}

An overview of our method, shown in Figure \ref{fig:overview}, is as
follows. First, we take a dataset of free-space demos and augment them
with randomly generated arcs and record the augmentation parameters
as a conditioning variable. We then train a flow-matching policy to
recreate these trajectories, with the augmentation label injected
into the model via AdaLN-Zero \cite{peebles2023dit}. At inference we
sample a batch of $K$ plans with different randomly sampled labels from
the training support, score them with an analytic collision cost,
and execute the best trajectory. This allows us to get reliable
zero-shot obstacle avoidance, without train-time knowledge of the
obstacle geometry or assumptions about the coverage of the prior.

The hyperparameters used across all experiments are listed in Table
\ref{tab:hparams}.

\begin{table}[t]
  \caption{DetAug hyperparameters. Hardware values given in
  parentheses where they differ.}
  \label{tab:hparams}
  \centering
  \small
  \begin{tabular}{ll}
    \toprule
    Parameter & Value\\
    \midrule
    \multicolumn{2}{l}{\textit{Augmentation}}\\
    Arc half-angle $\varphi$ & $[0.15\pi,0.85\pi]$ ($[0.25\pi,0.6\pi]$)\\
    Arc plane rotation $\theta$ & $[0,\pi]$\\
    Transit-phase margin & 0.4 s (1.5 s)\\
    Augmented copies per demo & 8 (6)\\
    \midrule
    \multicolumn{2}{l}{\textit{Architecture}}\\
    Total parameters & 10M\\
    Layers & 8\\
    Hidden dimension & 256\\
    Attention heads & 4\\
    Horizon $H$ & 50\\
    Action steps $N$ & 10 (20)\\
    \midrule
    \multicolumn{2}{l}{\textit{Inference}}\\
    Sampled labels $K$ & 32 (8)\\
    Integration steps & 10\\
    \bottomrule
  \end{tabular}
\end{table}

\subsection{Architecture}

We use a DiT-style \cite{peebles2023dit} transformer as our policy
architecture, training with a conditional flow matching
\cite{lipman2023flow} loss. Our model outputs a chunk of $H$ states
and actions $\tau = [s_{1:H}, a_{1:H}]$, and we replan every $N$
action steps like a standard action chunking model
\cite{zhao2023aloha, chi2023diffusion}. Training windows are sampled
at random positions within each demonstration, with windows that run
past the end of an episode padded with its final state. The
conditioning vector $c = \phi_t(t) + \phi_z(z)$ is injected through
AdaLN-Zero \cite{peebles2023dit} conditioning, where $t$ is the
flow-matching time and $z$ the augmentation label, with $\phi$ being
their respective networks. This is a common choice of conditioning method
for DiT models, but we also found it makes it easy to graft our
augmentation label onto an existing pre-trained policy, a setting we
test in Section \ref{sec:finetune}.

\subsection{Detour Augmentation}

\begin{figure}[t]
  \centering
  \includegraphics[width=\columnwidth]{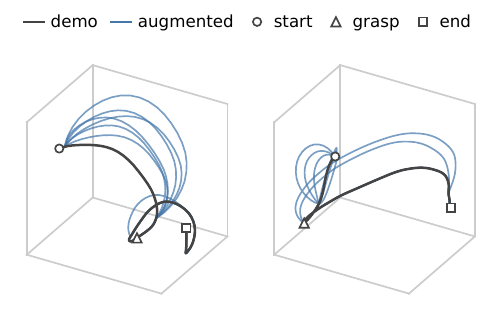}
  \caption{Example detour augmentations. Each panel shows an original
    end-effector demonstration and several augmented variants, with the
    approach and carry phases replaced by arcs of varying half-angle
    $\varphi$ and plane rotation $\theta$, while the grasp and release
  windows are left unchanged.}
  \label{fig:trajectories}
\end{figure}

To generate augmented trajectories, we first split demonstrations into
interaction and transit phases, similar to the motion and skill
stages of DemoGen \cite{xue2025demogen}. In our case interaction
phases are a fixed window around each gripper transition and transit
phases are the motion between them.
Each grasp yields two transit phases: an \textit{approach} from the
start to the grasp, and a \textit{carry} from the grasp to the
release. We replace each of these, minus some margin on either side,
with a sampled arc. These are generated by replacing the end-effector
chord between the endpoints with a circular arc of half-angle
$\varphi\in[\varphi_{\min},\varphi_{\max}]$ in a plane rotated by
$\theta\in[0,\pi]$ about the chord. The label for an arc is
$\varphi(\cos\theta,\sin\theta)\in\mathbb{R}^2$, so a grasp
carries a 4-dimensional label $z\in\mathbb{R}^4$ formed by
concatenating the approach and carry labels. This gives us a smooth,
continuous parameterisation that is zero for unbent demos. We
then check samples for kinematic feasibility and joint motion
smoothness, rejecting any samples that fail these criteria. For tasks
with multiple grasps like some in the SafeLIBERO
benchmark in Section \ref{sec:sim}, we augment each segment separately and
concatenate the labels into a single descriptor. Figure
\ref{fig:trajectories} shows example augmentations produced by this
procedure: the arcs sweep out a family of detours around the original
chord while the grasp and release windows are preserved exactly.

\subsection{Inference-Time Selection}

To select a trajectory at inference time, we sample a batch of $K$
augmentation labels uniformly from within the training support, and
generate a trajectory for each sample, conditioned on the current
observation and optional goal state. This is repeated with a fresh
batch at every replan rather than latching the first choice; we found
latching roughly halved collision-free success, as an early pick made
under a partial view of the scene is rarely the best one later. We
then score the trajectories with a collision cost, computed using
points sampled along the robot's surface against a signed distance
field from the obstacle. Importantly, we score trajectories by penetration only,
rather than by clearance. We found the latter tended to
produce behaviour that would ``game'' the metric via overly circuitous
paths and often hurt task success.

\section{EXPERIMENTAL RESULTS}

\subsection{SafeLIBERO Benchmark}\label{sec:sim}

In order to get a broad and diverse benchmark of novel obstacle avoidance
capabilities, we first run a series of experiments on the simulated
SafeLIBERO \cite{hu2025vlsa} benchmark. The base dataset used for
training contains the scenes from the standard LIBERO \cite{liu2023libero}
benchmark, but at test time, random obstacles are inserted into the
scene with varying levels of difficulty. For comparison, we try to
pick a representative method from all of the alternative approaches
mentioned previously. DemoGen \cite{xue2025demogen} represents a ``train-time
oracle'' with access to the obstacle geometries at training time; AEGIS
\cite{hu2025vlsa} is the method from the original SafeLIBERO paper
and is a purely inference-time technique, where a CBF-QP is applied
to the output of a frozen policy; CAPE \cite{yang2025cape} represents another
inference-time method, applying collision guidance to a re-noised
copy of the unexecuted plan.

For quick iteration and in order to
highlight just the contribution of the obstacle avoidance method, we
use a state-based policy with access to sim states instead of
vision. All of these techniques can scale to vision-based methods,
with view synthesis from augmented demos following the method from
DemoGen \cite{xue2025demogen}, but their relative performances would
likely be unchanged, as prior benchmarks report broadly consistent
method rankings under state and image observations \cite{mandlekar2021robomimic,
chi2023diffusion}. The
collision-free success rate, the overall success rate, and the
collision avoidance rate for each of the test suites, averaged
over 3 seeds, are reported in Table \ref{tab:sim_results}. An
important detail to note is that
we found several of the tasks in the benchmark to be infeasible, due
to the fact that there were no object interaction segments in the dataset
that would not be in collision with the obstacles. As all methods
scored close to $0\%$ success on these, we chose to omit them
entirely from the benchmark. In total this represented about 20\%
of the tasks.

DetAug achieved a 52.2\% clean success rate, more than 20pp
higher than the next best method. Its advantage over the two families
of methods is reflected in how this breaks down. The inference-time
methods AEGIS and CAPE have high avoidance rates, with CAPE the best
on average, but low success, and often end in timeouts as the policy
is unable to find a feasible path. Conversely, the train-time method
DemoGen, even though it has access to the obstacle geometry at
test time, is conditioned on a single max-pooled point cloud
embedding inherited from its DP3 backbone \cite{ze2024dp3}. We
found this was too weak a conditioning signal to push the
trajectories along significantly different paths, hence the low avoidance rate.

\subsection{Adapting a Pre-Trained Policy}\label{sec:finetune}

\begin{figure}[t]
  \centering
  \includegraphics[width=\columnwidth]{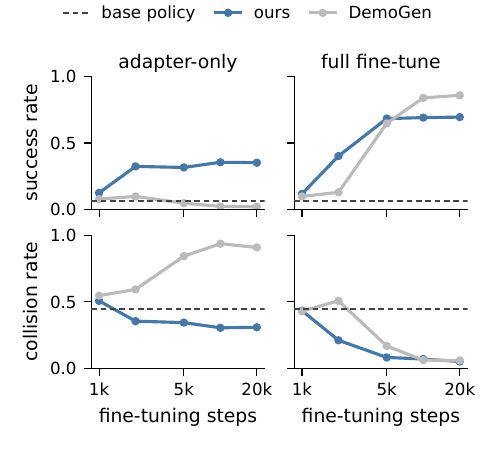}
  \caption{Success and collision rates over fine-tuning steps when
    adapting a pre-trained base policy with DetAug (ours) and DemoGen,
  training either the AdaLN adapter only or the full policy.}
  \label{fig:finetune}
\end{figure}

\begin{figure}[t]
  \centering
  \includegraphics[width=\columnwidth]{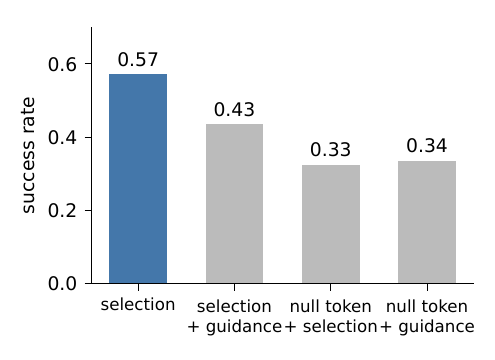}
  \caption{Collision-free success on the SafeLIBERO spatial suite for
  DetAug's label selection against guidance and mixed variants.}
  \label{fig:selection}
\end{figure}

\begin{figure*}[t]
  \centering
  \includegraphics[width=\textwidth]{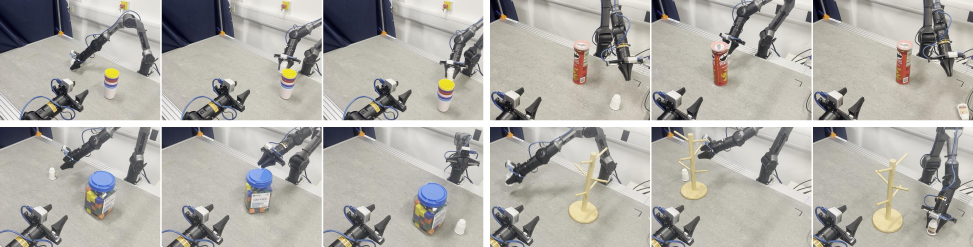}
  \caption{Hardware experiments. DetAug rollouts on the pick-and-place
    task with the four test-time obstacles: cups (top left), tube (top
    right), jar (bottom left) and rack (bottom right). The policy was
    trained on free-space demos only, and obstacles are
  placed at random positions with feasible solutions.}
  \label{fig:hardware}
\end{figure*}

\begin{figure}[t]
  \centering
  \includegraphics[width=\columnwidth]{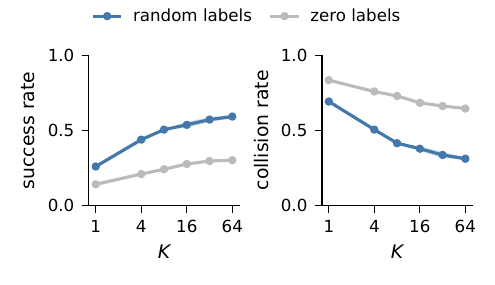}
  \caption{Collision-free success on the SafeLIBERO spatial suite as
    the number of sampled labels $K$ varies, for random and zero-valued
  labels.}
  \label{fig:ncond}
\end{figure}

We also test the performance of our method in a fine-tuned setting,
where we want to add obstacle avoidance capabilities to an existing
policy. To do this, we simulate a pre-existing policy by using the
same state-based architecture described in Section \ref{sec:method},
but just trained on a base, unaugmented dataset. We then compare it
to DemoGen \cite{xue2025demogen} by using AdaLN \cite{peebles2023dit}
to inject the conditioning, i.e. a point cloud embedding for DemoGen
and the augmentation label for DetAug. As we found DemoGen's capacity
was insufficient for the full range of obstacles in SafeLIBERO, we
restrict the comparison to a single pick-and-place task in the Newton
simulator \cite{newton2025}, where a single cylindrical obstacle is
added at test time. We test two different settings: \textit{adapter
only}, where we train the label encoder $\phi_z$ and the AdaLN
modulation layers while keeping the rest of the backbone frozen,
about 3M parameters in total, and \textit{full fine-tune}, where we
also train the main policy. The results for this are shown in Figure
\ref{fig:finetune}.

In the adapter-only setting, our model far outperforms DemoGen,
reaching a final success rate of 35\% versus DemoGen's 2\%, which is
even worse than the base policy's 6\%. The collision rates tell a
similar story, with DemoGen degrading well above the base rate.
However, full fine-tuning yields a very different outcome, where
DemoGen is actually able to beat DetAug in success rate and match us
in collision rate, albeit with a greater number of training steps. We
conclude that in a limited-compute setting, simple augmentation label
conditioning is far more capable at adding obstacle avoidance
capabilities to an existing policy than point cloud conditioning.
However, if full fine-tuning is possible, point cloud conditioning can
match or even outperform our method. This however comes with a steep
cost, as training a VLA with billions of parameters can be
prohibitively expensive for many users, and even when it is not, this process
can still take many days to fully converge.

\subsection{Selection vs Guidance}

To justify our inference-time selection method, we run another
experiment on the SafeLIBERO spatial suite. In particular, we aim to
answer the question \textit{How does our selector compare to using
guidance?}, which is a common technique for obstacle avoidance used in
methods like CAPE \cite{yang2025cape}. To do this, we train our
policy with the augmentation label randomly dropped out for a ``null
token'' \cite{ho2022cfg}, allowing us to generate unconditional
samples. In this
setting, we can then use the gradient of the same collision method
used by DetAug as a guidance function in an attempt to generate
collision-free samples without selection. The results of this experiment
are shown in Figure \ref{fig:selection}. \textbf{Null token + guidance} is the
method just described and performs significantly worse than
DetAug, 26pp lower. The two other columns show methods that are a mix
of these two approaches, either using guidance and selection, or
selecting over unconditional null token samples. Both of these mixes
also perform worse. We hypothesise two mechanisms behind this.
Firstly, the augmentation label \textit{reduces the dimensionality of
the search space}. Searching over the 4-dimensional label space is
much easier than the full 52-dimensional state and action space,
increasing the likelihood of finding collision-free samples.
Secondly, the augmentation label represents the parameters of the
\textit{clean} sample, whereas guidance is applied to intermediate
\textit{noisy} samples, which are by definition off-manifold. As a
result, the dynamics of the final generated sample are much more likely to
be incorrect.

\subsection{Effect of the Number of Sampled Labels}\label{sec:ncond}

Our final ablation aims to identify the effect of the number of
sampled labels at inference time on policy performance. To do this we
again roll out our policy in the SafeLIBERO spatial suite, but vary
the number of samples $K$ between 1 and 64. As a control, we also evaluate
the performance with different numbers of zero-valued labels,
corresponding to unaugmented demos. The results are shown in Figure
\ref{fig:ncond}. There are two main takeaways. Firstly, for both random and
zero labels, the success rate increases monotonically, as expected.
Secondly, and more importantly, random labels are strictly
better than zeros, further validating our method. This experiment
also explains why we chose $K=32$ for our experiments; $K=64$ does
give a success rate about $2\%$ higher, but at the expense of about
double the inference cost.

\subsection{Hardware Experiments}\label{sec:hardware}

\begin{table}[t]
  \caption{Hardware success rates (\%), 10 rollouts per obstacle
  type. The best result in each column is shown in bold.}
  \label{tab:hardware}
  \centering
  \small
  \begin{tabular}{l|cccc|c}
    \toprule
    Method & Cups & Tube & Jar & Rack & \textbf{Average}\\
    \midrule
    DetAug (ours) & \textbf{90} & \textbf{90} & \textbf{70} &
    \textbf{100} & \textbf{87.5}\\
    DemoGen~\cite{xue2025demogen} & \textbf{90} & \textbf{90} & 50 & 90 & 80\\
    CAPE~\cite{yang2025cape} & 50 & 30 & 30 & 30 & 35\\
    AEGIS~\cite{hu2025vlsa} & 50 & 40 & 20 & 10 & 30\\
    \bottomrule
  \end{tabular}
\end{table}

We lastly verify the validity of our method on hardware by first
training a policy on a simple pick-and-place task with free-space
demonstrations,
and then rolling out each method with four obstacles of different
shapes placed in random positions with feasible solutions as shown in Figure
\ref{fig:hardware}. We made sure to keep the distribution of
positions as close as possible between trials. We compare DetAug to
the same three other methods used in the simulation benchmark. The
results are listed in Table \ref{tab:hardware}. DetAug is the best
performing method at 87.5\%, with DemoGen second at 80\%. DemoGen's
failures mainly came from moving straight through the
obstacle in scenarios where large detours were required. The clear
result is the collapse of the two inference-time methods to 35\% and
30\% respectively. Both had similar drawbacks: the policies would
struggle with significant detours from the prior, and because neither
separates grasping from transit, the corrections would interfere with
the object interaction, an issue DetAug avoids by construction since
the grasp and release windows are never modified.

\section{CONCLUSIONS}

In summary, we present DetAug, a method for adapting manipulation
policies with zero-shot obstacle avoidance that does not require any
knowledge of the test-time environment during training. We show that
it significantly outperforms prior methods on a broad suite of
simulation tasks and that selection over the augmentation label
outperforms guidance on the same policy. We demonstrate a superior
ability to adapt pre-trained policies with a limited compute budget
compared to methods conditioned on the obstacle geometry, and
finally verify its performance on real hardware tasks, where
inference-time methods collapse on large detours and DetAug matches
or exceeds an obstacle-conditioned baseline that saw obstacles in
training. In future work, we would be interested
in scaling this method to a large VLA model and to tasks requiring
greater dexterity.

%%%%%%%%%%%%%%%%%%%%%%%%%%%%%%%%%%%%%%%%%%%%%%%%%%%%%%%%%%%%%%%%%%%%%%%%%%%%%%%%

%%%%%%%%%%%%%%%%%%%%%%%%%%%%%%%%%%%%%%%%%%%%%%%%%%%%%%%%%%%%%%%%%%%%%%%%%%%%%%%%

%%%%%%%%%%%%%%%%%%%%%%%%%%%%%%%%%%%%%%%%%%%%%%%%%%%%%%%%%%%%%%%%%%%%%%%%%%%%%%%%

\section*{ACKNOWLEDGMENT}

Generative AI tools (Claude, Anthropic) were used throughout the
preparation of this work, including drafting and editing text in all
sections, generating and debugging code for experiments and figures,
and assisting with the analysis of results. All AI-generated content
was reviewed and verified by the authors, who take full
responsibility for the final manuscript.

%%%%%%%%%%%%%%%%%%%%%%%%%%%%%%%%%%%%%%%%%%%%%%%%%%%%%%%%%%%%%%%%%%%%%%%%%%%%%%%%

% Balance the columns on the last page
\IEEEtriggeratref{25}
\bibliographystyle{IEEEtran}
\bibliography{references}

\end{document}